%% file: acl_latex.tex
\pdfoutput=1

\documentclass[11pt]{article}

\usepackage[final]{acl}

\usepackage{times}
\usepackage{latexsym}

\usepackage[T1]{fontenc}

\usepackage[utf8]{inputenc}

\usepackage{microtype}

\usepackage{inconsolata}

\usepackage{graphicx}
\usepackage{amssymb}
\usepackage{bm}
\usepackage{amsmath}
\usepackage{algorithm}
\usepackage{algorithmic}
\usepackage{booktabs}
\usepackage{multirow}
\usepackage{multicol}
\usepackage{pifont}
\usepackage{enumitem}
\usepackage{colortbl}
\usepackage{subcaption}

\definecolor{myred}{RGB}{0, 0, 255}

\title{Embracing Large Language Models in Traffic Flow Forecasting}

\makeatletter
\renewcommand{\@fnsymbol}[1]{^\dagger}
\makeatother

\author{
Yusheng Zhao\textsuperscript{$\heartsuit$}, Xiao Luo\textsuperscript{\ding{171}\textdagger}, Haomin Wen\textsuperscript{$\diamondsuit$}, Zhiping Xiao\textsuperscript{\ding{168}\textdagger}, Wei Ju\textsuperscript{$\heartsuit$}, Ming Zhang\textsuperscript{$\heartsuit$\textdagger} \\
\textsuperscript{$\heartsuit$} State Key Laboratory for Multimedia Information Processing,\\ School of Computer Science, PKU-Anker LLM Lab, Peking University \\ \textsuperscript{\ding{171}} Department of Computer Science, University of California, Los Angeles \\
\textsuperscript{$\diamondsuit$} Carnegie Mellon University \\ \textsuperscript{\ding{168}} Paul G. Allen School of Computer Science and Engineering, University of Washington \\
\texttt{yusheng.zhao@stu.pku.edu.cn}, 
\texttt{xiaoluo@cs.ucla.edu},\\
\texttt{wenhaomin.whm@gmail.com},
\texttt{patxiao@uw.edu}, 
\texttt{\{juwei,mzhang\_cs\}@pku.edu.cn}
}

\makeatletter
\def\blfootnote{\xdef\@thefnmark{}\@footnotetext}
\makeatother

\begin{document}
\def\method{LEAF}
\maketitle
\begin{abstract}
\input{1_abstract}
\end{abstract}
\blfootnote{
\textsuperscript{\textdagger} Corresponding authors.
}

\input{2_intro}
\input{3_related}

\input{4_method}
\input{5_experiment}
\input{6_conclusion}

\bibliography{custom}

\clearpage

\end{document}

%% file: 1_abstract.tex
Traffic flow forecasting aims to predict future traffic flows based on historical traffic conditions and the road network. It is an important problem in intelligent transportation systems, with a plethora of methods being proposed. Existing efforts mainly focus on capturing and utilizing spatio-temporal dependencies to predict future traffic flows. Though promising, they fall short in adapting to test-time environmental changes in traffic conditions. To tackle this challenge, we propose to introduce large language models (LLMs) to help traffic flow forecasting and design a novel method named Large \underline{L}anguage Model \underline{E}nhanced Tr\underline{a}ffic \underline{F}low Predictor (\method{}). \method{} adopts two branches, capturing different spatio-temporal relations using graph and hypergraph structures, respectively. The two branches are first pre-trained individually, and during test time, they yield different predictions. Based on these predictions, a large language model is used to select the most likely result. Then, a ranking loss is applied as the learning objective to enhance the prediction ability of the two branches. Extensive experiments on several datasets demonstrate the effectiveness of  \method{}. Our code is available at \url{https://github.com/YushengZhao/LEAF}.

%% file: 2_intro.tex
\section{Introduction}
Traffic flow forecasting is an integral part of intelligent transportation systems \cite{dimitrakopoulos2010intelligent, zhang2011data} and smart cities \cite{shahid2021towards, dai2022st}. The target of traffic flow forecasting is to predict future traffic flows using historical data and spatial information (\emph{i.e.} the road network), which has a wide range of applications including traffic signal control \cite{jiang2021urban}, route planning \cite{liebig2017dynamic}, and congestion management \cite{fouladgar2017scalable}. 

\begin{figure}
    \centering
    \includegraphics[width=\linewidth]{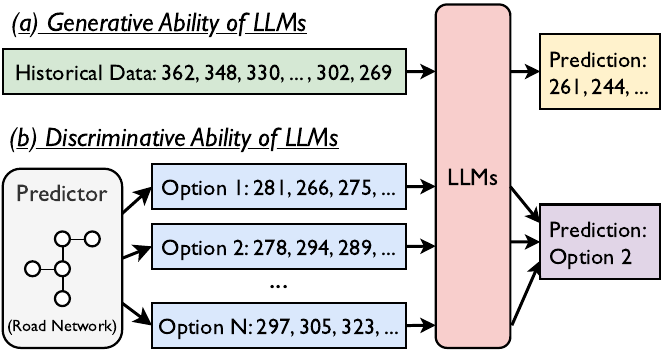}
    \caption{To use LLMs for traffic flow forecasting, a naive solution is to utilize their \textit{generative} ability (a), which is hard to incorporate spatial relations. By comparison, \method{} utilizes the discriminative ability of LLMs (b), making it easier for LLMs to predict.}
    \label{fig:motivation}
    
\end{figure}

Due to its value in real-world applications, great efforts have been made to resolve the problem of traffic flow forecasting \cite{smith1997traffic, sun2006bayesian, guo2019attention, li2021spatial}. Early works mainly model the traffic systems using physical rules or shallow models \cite{ghosh2009multivariate, tchrakian2011real, hong2011hybridSVR, li2012k}. With the advent of deep learning, the main-stream of traffic flow forecasting methods utilizes graph neural networks \cite{kipf2016semi, hamilton2017inductive, xu2018powerful, velivckovic2018graph}, recurrent neural networks \cite{hochreiter1997long, chung2014empirical}, and transformers \cite{vaswani2017attention, jiang2023pdformer} to capture the rich spatio-temporal relations \cite{yu2018spatioSTGCN, li2018diffusionDCRNN, guo2019attention, li2021spatial, zhang2021traffic, chen2022bidirectional, liu2023spatioSTAEformer, ma2024spatio}.

Despite their success, existing traffic flow forecasting methods have two limitations, which hinder their applications in the real world. 

\textit{Firstly, existing methods are generally unable to adapt to environmental changes of traffic conditions during test time.} Most existing methods make the assumption that test data follow the same distribution as the training data, which may fail to hold in the real world \cite{lu2022out, zhao2025attention, zhao2025traci}, especially for time series data \cite{kim2021reversible, fan2023dish, chen2024calibration, zhang2024fast}. Traffic conditions change over time due to a variety of factors like special events, changes in weather, or the shift of eras. Traditional models \cite{song2020spatialSTSGCN, li2021spatial}, including fusion models \cite{kashinath2021review} have poor generalization ability, suffering from performance degradation under distribution shifts. In contrast, LLMs have \textbf{strong generalization ability} to adapt well under such distribution shifts \cite{chang2024survey, minaee2024large, beniwal2024remember}. Moreover, LLMs have \textbf{strong reasoning ability}, which enables them to infer the current environment from the data to benefit forecasting. Due to the \textbf{high cost of collecting large-scale data}, the zero-shot generalization/reasoning ability is especially useful.
To utilize LLMs, a naive solution is to use the \textit{generative} ability of LLMs to make direct predictions \cite{li2024urbangpt, ren2024tpllm, liang2024exploring}, as shown in Figure \ref{fig:motivation} (upper part). However, this is too challenging for language models, as accurate forecasting relies on both historical data and complex spatio-temporal relations. 

\textit{Secondly, existing methods are weak in capturing the rich structure of spatio-temporal relations in traffic data.} The traffic network is complicated and the temporal dimension adds another layer of complexity. Prior works focus on capturing the complex spatio-temporal relations, using graph structures \cite{song2020spatialSTSGCN, zheng2023spatio} or hypergraph structures \cite{wang2022multitask, wang2022shgcn, zhao2023dynamic}. Graphs capture pair-wise relations, while hypergraphs model non-pair-wise relations. Adopting only one of them is not enough, as the spatio-temporal relations in traffic data are rich by nature. For example, traffic congestion at one vertex affects adjacent vertices (pair-wise relations), whereas road closures affect a large set of vertices (non-pair-wise relations). Modeling the rich structure of spatio-temporal relations is challenging in predicting future traffic flows.

To that end, this paper proposes a novel method termed Large \underline{L}anguage Model \underline{E}nhanced Tr\underline{a}ffic \underline{F}low Predictor (\method{}) for adaptive and structure-perspective traffic flow forecasting. 
The core idea of \method{} is to utilize the discriminative ability of LLMs to enhance the task of traffic flow forecasting using a predictor and a selector, where the predictor generates predictions and the selector chooses the most likely result. To enhance adaptability, we build an LLM-based selector that selects from a range of possible future traffic flows using the discriminative ability of an LLM, as shown in Figure \ref{fig:motivation} (lower part). The selection results are used to guide the predictor with a ranking loss \cite{weinberger2009distance, sohn2016improved}, which only requires that the positive candidate (the ones chosen by the LLM) is better than the negative candidates (the ones not chosen). The LLM is good at understanding the changing traffic conditions and is open to further information provided by humans, making it an adaptable predictor. To better capture the rich structures of spatio-temporal relations, we build a dual-branch predictor composed of a graph branch which captures pair-wise relations of spatio-temporal traffic data, and a hypergraph branch, which captures non-pair-wise relations. During test time, the dual-branch predictor generates different forecasting results, and subsequently, a set of transformations is applied to obtain a wealth of choices of future traffic flows.

Our contribution is summarized as follows:
\begin{itemize}[itemsep=0mm,left=2mm, topsep=1mm]
    \item We propose an LLM-enhanced traffic flow forecasting framework that introduces LLMs in test time to enhance the adaptability under distribution shift of traffic flow forecasting models.
    \item We propose a dual-branch predictor that captures both pair-wise and non-pair-wise relations of spatio-temporal traffic data, and an LLM-based selector that chooses from possible prediction results generated by the predictor. The selection results further guide the adaptation of the predictor with a ranking loss.
    \item Extensive experiments on several datasets verify the effectiveness of the proposed method.
\end{itemize}

%% file: 3_related.tex
\section{Related Works}
\subsection{Traffic Flow Forecasting}
Traffic flow forecasting is a topic that has been studied for several decades \cite{smith1997traffic, sun2006bayesian, yang2016optimized, song2020spatialSTSGCN, li2024urbangpt, li2024graph}. Early efforts mainly focus on traditional models \cite{smith1997traffic, asadi2012new}. With the success of deep learning, deep neural networks have become mainstream in this field. One line of research adopts recurrent neural networks (RNNs) \cite{hochreiter1997long} and graph neural networks (GNNs) \cite{kipf2016semi}, where the GNNs and RNNs capture the spatial and temporal relations, respectively \cite{li2018diffusionDCRNN, wang2020traffic, chen2022aargnn, li2023dynamic, weng2023decomposition}. 

To jointly model spatial and temporal relations, another line of research utilizes GNNs in both dimensions \cite{song2020spatialSTSGCN, li2021spatial, lan2022dstagnn, chen2024traffic}. As simple graphs only capture pair-wise relations, some works take a step further, introducing hypergraph neural networks (HGNNs) to capture non-pair-wise spatio-temporal relations \cite{luo2022directed, wang2022shgcn, sun2022dual, zhao2023dynamic, wang2024large}. In this work, we take advantage of the benefits from both sides, designing a dual-branch architecture, which better captures the rich structures of spatio-temporal relations.

\begin{figure*}
    \centering
    \includegraphics[width=\linewidth]{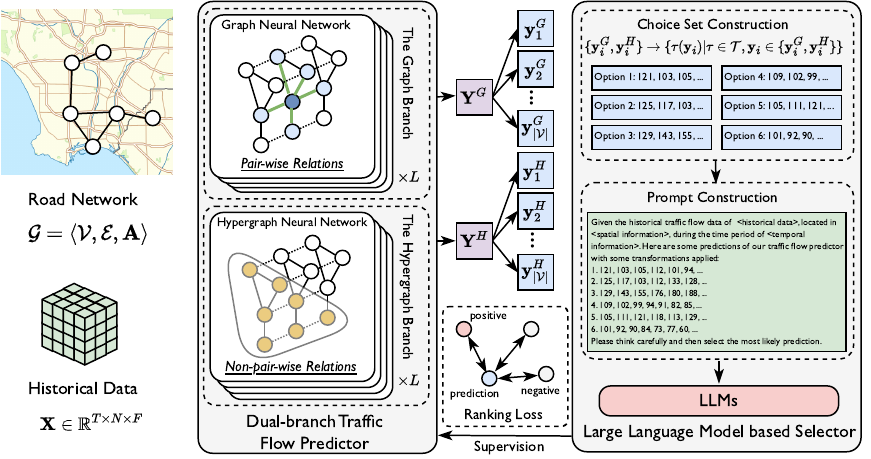}
    \caption{The framework of the proposed \method{}, consisting of a dual-branch predictor and an LLM-based selector. The predictor generates forecasts of traffic flows with the graph and hypergraph branches. The selector constructs a choice set and selects the best option using an LLM. The selection results are used to supervise the predictor.}
    \label{fig:framework}
    \vspace{-1.5mm}
\end{figure*}

\subsection{Large Language Models}
In recent years, large language models have drawn increased attention, both within the community of natural language processing \cite{chen2024large, yin2024survey, luo2025large} and beyond, including healthcare \cite{lievin2024can, van2024adapted, labrak2024biomistral}, education \cite{milano2023large}, legal technology \cite{lai2024large}, economics \cite{li2024econagent}, recommendation \cite{chen2024large, bao2024large}, code understanding \cite{zhao2025marco}, and transportation \cite{liu2024spatial, guo2024explainable, ren2024tpllm}.

Among these applications, traffic flow forecasting is an important one, as it serves as the foundation of many downstream tasks in the field of intelligent transportation systems \cite{boukerche2020artificial}. The success of LLMs in language has inspired a plethora of works in this task. Some works adopt the architecture of LLMs for building transformer-based traffic flow predictors \cite{cai2020traffic, xu2020spatial, chen2022bidirectional, liu2023spatioSTAEformer, jiang2023pdformer, zou2024multispans}. However, they typically require a large amount of data for training. 

Another line of research attempts to equip LLMs with the ability to predict future traffic flows based on the history and specific situations \cite{zheng2023chatgpt, guo2024towards, li2024urbangpt, yuan2024unist, han2024event}. Although LLMs have shown promising results in understanding time series \cite{yu2023harnessing, jin2023time, gruver2024large, koval2024financial, gruver2024large, ansari2024chronos, liu2024timer, woo2402unified} or temporal events \cite{xia2024chain, hu2024moments}, the traffic data involves complex spatio-temporal relations challenging LLMs' \textit{generative} ability. In this work, we show that one can instead utilize their \textit{discriminative} ability to enhance existing traffic flow forecasting models.

%% file: 4_method.tex
\section{Methodology}

\noindent\textbf{Problem Setup}. We follow the standard setup for traffic flow forecasting \cite{song2020spatialSTSGCN}, where there is a road network denoted as $\mathcal G=\langle \mathcal V, \mathcal E, \bm A \rangle$. $\mathcal V$ denotes the set of $N$ vertices (\emph{i.e.} the sensors in the city), $\mathcal E$ denotes the set of edges, and $\bm A\in \mathbb R^{N\times N}$ represents the adjacency matrix. In this road network, historical traffic flows can be represented as $\bm X=(\bm X_1, \bm X_2, \cdots, \bm X_{T})\in \mathbb R^{T\times N\times F}$, in which $T$ is the length of historical observation and $F$ is the dimension of input features. The goal is to predict the future of traffic flows with the length of $T'$, denoted as $\bm X'=(\bm X'_1, \bm X'_2, \cdots, \bm X'_{T'})\in \mathbb R^{T'\times N \times F}$.

\smallskip
\noindent\textbf{Framework Overview}. To solve the problem with the help of LLMs, we propose a novel framework termed Large Language Model Enhanced Traffic Flow Predictor (\method{}), whose overview is illustrated in Figure \ref{fig:framework}. Specifically, to achieve LLM-enhanced prediction (Section \ref{sec:prediction}), we design a dual-branch traffic flow predictor (Section \ref{sec:predictor}) and an LLM-based selector (Section \ref{sec:selector}). The predictor consists of a graph neural network branch capturing pair-wise spatio-temporal relations, and a hypergraph neural network branch capturing non-pair-wise relations. During training, the predictor is first pretrained using the training data. During test time, we apply transformations to the forecasting results of the predictor to obtain a variety of choices, among which the selector chooses the best one with a frozen LLM. The selection results are then used to fine-tune the dual-branch predictor.

\subsection{Dual-branch Traffic Flow Predictor}
\label{sec:predictor}
Previous works on traffic flow forecasting adopt \textit{the graph perspective} \cite{song2020spatialSTSGCN, zheng2023spatio} or \textit{the hypergraph perspective} \cite{wang2022multitask, zhao2023dynamic}. The graph perspective propagates messages between pairs of nodes, which makes them adept in modeling pair-wise spatio-temporal relations (\emph{e.g.} an accident affects adjacent locations). On the other hand, the hypergraph perspective propagates messages among groups of nodes, making them proficient in modeling non-pair-wise spatio-temporal relations (\emph{e.g.} people move from the residential area to the business area). For \method{}, we aim to take the benefits from both sides by constructing a dual-branch predictor and letting the LLM select.

\smallskip
\noindent\textbf{Spatio-temporal Graph Construction}. To utilize graph neural networks and hypergraph neural networks, we first construct a spatio-temporal graph corresponding to the input tensor $\bm X \in \mathbb R^{T\times N\times F}$. Particularly, the length of historical data $T$ yields a set of $TN$ spatio-temporal nodes, denoted as:
\begin{equation}
    \mathcal V^{ST}=\left\{v_i^t\mid i=1,\dots,N, t=1,\dots,T\right\},
\end{equation}
and we add temporal edges in addition to spatial edges $\mathcal E$ to obtain the edge set $\mathcal E^{ST}$:
\begin{align}
\label{eq:st-edge}
\mathcal{E}^{ST} = \bigl\{ \langle v_i^{t_1}, v_j^{t_2} \rangle \mid \left( |t_1 - t_2| = 1 \land i = j \right) \nonumber \\
\quad \lor \left( t_1 = t_2 \land \langle i, j \rangle \in \mathcal{E} \right) \bigr\}.
\end{align}
The spatio-temporal graph can then be represented as $\mathcal G^{ST}=\langle \mathcal V^{ST},\mathcal E^{ST}, \bm A^{ST}\rangle$, where $\bm A^{ST}\in \mathbb R^{TN\times TN}$. The spatio-temporal features $\bm X^{(0)}\in \mathbb R^{TN\times d}$ is derived from $\bm X\in \mathbb R^{T\times N\times F}$ with a linear mapping and a reshape operation, where $d$ is the dimension of the hidden space.

\smallskip
\noindent\textbf{The Graph Branch}. Based on the constructed spatio-temporal graph, we first adopt a graph neural network to model pair-wise spatio-temporal relations. Concretely, given the spatio-temporal feature inputs $\bm X^{(0)}\in \mathbb R^{TN\times d}$, we adopt convolution layers to process the features, which is 
\begin{equation}
\label{eq:gconv}
\bm X^{(l)}=\sigma\left(\widehat{\bm {A}^{ST}}\bm X^{(l-1)}\bm W_G^{(l)}\right),
\end{equation}
where $\sigma(\cdot)$ is the activation layer and $\bm W_G^{(l)}\in \mathbb R^{d\times d}$ is the weight matrix. $\widehat{\bm A^{ST}}=\bm D^{-1/2}\bm A^{ST}\bm D^{-1/2}$ denotes the normalized version of the adjacency matrix, where $\bm D$ is the degree matrix \cite{kipf2016semi, song2020spatialSTSGCN}. By adopting graph convolutions in Eq. \ref{eq:gconv}, the information from one spatio-temporal vertex can be propagated to its neighbors as defined in $\mathcal E^{ST}$ in Eq. \ref{eq:st-edge}, and thus this branch models pair-wise spatio-temporal relations.

\smallskip
\noindent\textbf{The Hypergraph Branch}. Although the graph branch is adept at capturing pair-wise relations, the complex traffic patterns contain non-pair-wise relations. For example, in the morning rush hours, people move from the residential area (which is a set of vertices in a hyperedge) to the business area (which is another hyperedge). The vertices in one hyperedge share common patterns and the hyperedges affect each other. To model non-pair-wise relations, hypergraphs are adopted. For a hypergraph, its incidence matrix $\bm I_H \in \mathbb R^{NT\times m}$ describes the assignment of $NT$ vertices to $m$ hyperedges. As the incidence matrix is not given as input, we resort to a learnable one with low-rank decomposition \cite{zhao2023dynamic}:
\begin{equation}
\bm I_H = \operatorname{softmax}(\bm X^{(l-1)}_H \bm W_H),
\end{equation}
where $\bm X^{(l-1)}_H\in \mathbb R^{NT\times d}$ is the hidden features, and $\bm W_H\in \mathbb R^{d\times m}$ is the weight matrix. $\operatorname{softmax}(\cdot)$ is applied for normalization. Subsequently, the output features can be computed as:
\begin{equation}
\label{eq:hgcn}
\bm X^{(l)}_H=\bm I_H\left(
\bm I_H^\top \bm X^{(l-1)}_H + 
\sigma\left(\bm W_E \bm I_H^\top \bm X^{(l-1)}_H\right) 
\right),
\end{equation}
where $\bm W_E\in \mathbb R^{m\times m}$ models the interactions of the hyperedges. In this way, the hypergraph branch considers both (\textit{a}) the interactions within a set of vertices (within a hyperedge) through the first term of Eq. \ref{eq:hgcn}, and (\textit{b}) the interactions among groups of vertices (among the hyperedges) through the second term of Eq. \ref{eq:hgcn}.

\begin{figure}
    \centering
    \includegraphics[width=\linewidth]{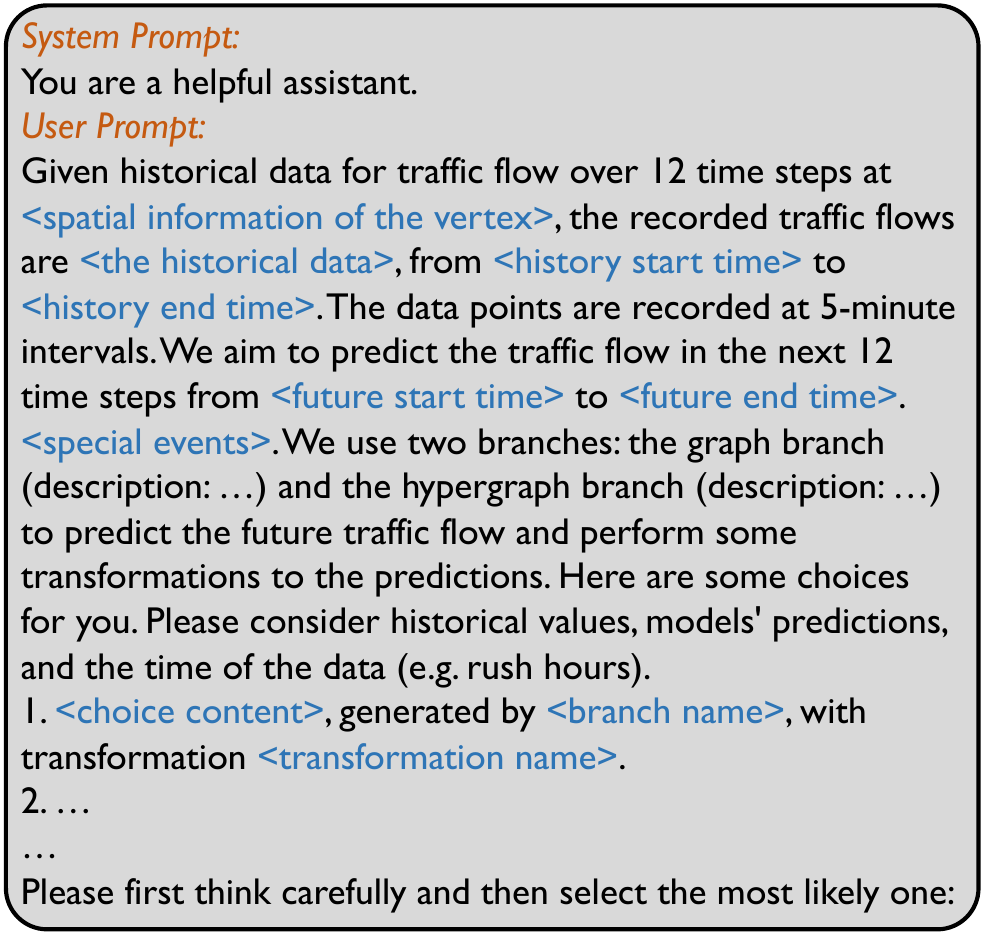}
    \caption{An illustration of the prompt template.}
    \label{fig:prompt}
    \vspace{-3mm}
\end{figure}
\subsection{LLM-Based Selector}
\label{sec:selector}

In Section \ref{sec:predictor}, we obtain different prediction results from different branches, denoted as $\bm Y^G$ and $\bm Y^H$. Most previous works \cite{li2024urbangpt, ren2024tpllm, liang2024exploring} directly generate these predictions, which is challenging since the complex spatio-temporal relations are hard to express in texts. By comparison, the LLM-based selector aims to choose the best prediction, using the internal knowledge of a frozen LLM and the constructed prompts. As the traffic networks are usually large, we break the result $\bm Y\in\{\bm Y^G,\bm Y^H\}\subset  \mathbb R^{T'\times N}$ for individual vertices, \emph{i.e.} $\bm Y=[\bm y_1,\bm y_2,\dots,\bm y_N]$, where $\bm y_i\in \mathbb R^{T'}$.

\smallskip
\noindent\textbf{Choice Set Construction}. In practice, we want to give the LLM-based selector more choices so that it has the potential to make better predictions. Therefore, we introduce several transformations: smoothing, upward trend, downward trend, overestimating, and underestimating. The set of transformations is denoted as $\mathcal T$. For vertex $i\in \mathcal V$, the choice set is determined as follows:
\begin{equation}\label{eq:choice}
\mathcal C_i = \left\{ \tau(\bm y_i) | \tau\in\mathcal T,\bm y_i\in \{\bm y_i^G,\bm y_i^H\} \right\} \cup \{\bm y_i^G,\bm y_i^H\}.
\end{equation}
By adopting transformations, the choice set is expanded and the selector has the potential to deal with more complex situations. For instance, if it believes that both of the branches underestimate the traffic flow on a Monday morning, the selector can choose an option with an upward trend.

\smallskip
\noindent\textbf{Prompt Construction}. When constructing the prompt, we consider the following aspects. (1) General information about the data, including the meaning of the numbers, the way traffic data are selected, etc. (2) The spatial information of the vertex, including the sensor ID, and the geometric location information. (3) The temporal information, including the time historical data are collected, the time we want to forecast, and whether there are special events. (4) The historical data. (5) The task instructions. (6) The choice set constructed in Eq. \ref{eq:choice}, including the names of specific branches and the description of augmentations. An illustration of this is shown in Figure \ref{fig:prompt}.

\subsection{LLM-Enhanced Prediction}
\label{sec:prediction}
The \method{} framework consists of a predictor and a selector. The predictor (the graph branch and the hypergraph branch) is pre-trained on the training set. During test time, the predictor first predicts, and the selector then selects. The selection results are used to supervise the predictor, and thus the two modules benefit from each other, achieving LLM-enhanced prediction. Concretely, given the input data $\bm X \in \mathbb R^{T\times N\times F}$, the predictor generates two forecasts, \emph{i.e.} $\bm Y^G$ and $\bm Y^H$. Subsequently, the selector constructs choice sets and uses the LLM to find the best option for each individual vertex. The selection results are denoted as $\hat{\bm y}_i$, $i\in\mathcal V$, which are then used to supervise the predictor. Conceivably, $\hat{\bm y}_i$ may not be the optimal choice in the choice set, and therefore, directly using MAE or MSE losses may lead to noises in the supervision signals. By comparison, the ranking loss \cite{weinberger2009distance, sohn2016improved} only requires that the positive candidate (the ones chosen by the LLM) is better than the negative candidates (the ones not chosen), described as follows:
\begin{equation}
\label{eq:metric}
\mathcal L^G = [\Delta(\bm y_i^G,\hat{\bm y}_i)-\inf_{\bm y_i'\in \mathcal C_i\setminus\{\hat{\bm y}_i\}} \Delta(\bm y_i^G,\bm y_i')+\epsilon]_+,
\end{equation}
where $[\cdot]_+$ is the hinge function, $\Delta(\cdot,\cdot)$ is a distance measure, and $\epsilon$ is the margin. Similarly, we can define the loss function $\mathcal L^H$ using $\bm y_i^H$. The final objective is written as:
\begin{equation}
\label{eq:loss}
\mathcal L = \mathcal L^G+\mathcal L^H.
\end{equation}
In Eq. \ref{eq:metric} and Eq. \ref{eq:loss}, we encourage the forecasts of the predictor (\emph{i.e.} $\bm y_i^G$ and $\bm y_i^H$) to be closer to the selected prediction (\emph{i.e.} $\hat{\bm y}_i$) than the closest one in suboptimal predictions (\emph{i.e.} $\mathcal C_i\setminus\{\hat{\bm y}_i\}$). Since the ground truth may not be covered by the choice set, this objective is better than directly minimizing the distance between the predictions and the selected forecast, as it allows the model to learn from a better choice compared to suboptimal choices.

By supervising the predictor, the two modules (\emph{i.e.} the predictor and the selector) benefit from each other. A better predictor yields better choices, which benefits the selector; better selection results provide better supervision signals for the predictor. Through the iteration of prediction and selection, we can achieve LLM-enhanced prediction. The \method{} framework during inference is summarized in Algorithm \ref{alg:main}. 

\begin{algorithm}[tb]
    \caption{The Algorithm of \method{}}
    \label{alg:main}
    \textbf{Requires}: The road network $\mathcal G=\langle \mathcal V, \mathcal E, \bm A\rangle$, historical data $\bm X$, the number of iterations $K$, the graph branch $\mathcal B^G$, and the hypergraph branch $\mathcal B^H$.\\
    \textbf{Ensures}: The forecasting result $\hat{\bm Y}$.
    
    \begin{algorithmic}[1] 
    \STATE Construct spatio-temporal graph $\mathcal G^{ST}$.
    \FOR{$j=1,2,...,K$}
    \STATE Compute the prediction of the graph branch $\mathcal B^G$ as $\bm Y^G$.
    \STATE Compute the prediction of the hypergraph branch $\mathcal B^H$ as $\bm Y^H$.
    \STATE Construct the choice set using Eq. \ref{eq:choice}.
    \STATE Use LLM to select the best option for each vertex as $\hat{\bm y}_i$ where $i\in \mathcal V$.
    \STATE Use $\hat{\bm y}_i$ to supervise the predictor ($\mathcal B^G$ and $\mathcal B^H$) with Eq. \ref{eq:metric} and Eq. \ref{eq:loss}.
    \ENDFOR
    \STATE Stack $\hat{\bm y}_i, i\in \mathcal V$ to obtain $\hat{\bm Y}$
    \end{algorithmic}
\end{algorithm}

%% file: 5_experiment.tex
\input{Tab/main_old}
\section{Experiments}
\subsection{Experimental Setup}
\noindent\textbf{Datasets}. We adopt three widely used datasets in traffic flow forecasting, including PEMS03, PEMS04, and PEMS08. The datasets are publicly available and collected by California Transportation Agencies (CalTrans) Performance Measurement Systems (PEMS) \footnote{https://pems.dot.ca.gov/}. 
These records come from sensors on the roads of various places in California, and they are counted every five minutes. 

\smallskip
\noindent\textbf{Evaluation Metrics}. We follow standard setting \cite{li2018diffusionDCRNN} that use one-hour historical data (\emph{i.e.} $T=12$ timesteps) to forecast one-hour future (\emph{i.e.} $T'=12$ timesteps). 
To signify the test-time distribution shift, we adopt small training sets of 10\% of data and another 10\% for validation. 
We choose a subset of non-overlapping slices in the test set.
We adopt three standard metrics for evaluation, i.e., Mean Absolute Error (MAE), Root Mean Squared Error (RMSE), and Mean Absolute Percentage Error (MAPE) \cite{song2020spatialSTSGCN}. 

\smallskip
\noindent\textbf{Baseline Methods}. We compare \method{} with a variety of baselines, including 
DCRNN \cite{li2018diffusionDCRNN}, ASTGCN \cite{guo2019attention}, STSGCN \cite{song2020spatialSTSGCN}, HGCN \cite{wang2022multitask}, DyHSL \cite{zhao2023dynamic}, STAEformer \cite{liu2023spatioSTAEformer}, COOL \cite{ju2024cool}, and LLM-MPE \cite{liang2024exploring}. Among these methods, DCRNN is a combination of GNN and RNN. ASTGCN, STSGCN, and COOL are based on GNNs, while HGCN and DyHSL are based on hypergraph neural networks. STAEformer uses the transformer architecture, while LLM-MPE uses LLMs to understand and predict traffic data with textual guidance. 

\smallskip
\noindent\textbf{Implementation Details}. For the dual-branch predictor, the number of layers $L$ for both branches is set to 7. We use linear mapping as the input embedding, and the hidden dimension $d$ is set to 64, which is also shared across all baselines. We also use a two-layer MLP to map the last hidden embedding to the output. 
In choice set construction,  smoothing is implemented with an average filter, upward/downward trend increases/decreases the traffic flow by 1\% to 12\% (12 timesteps, linearly increasing), overestimate/underestimate increases/decreases the traffic flow by 5\% (for all 12 timesteps). 
As for the loss function in Eq. \ref{eq:metric}, we adopt Huber distance for $\Delta(\cdot,\cdot)$ and the margin $\epsilon$ is set to 0. When training with this loss function, we update the parameters for $M$ iterations, where $M$ is set to 5.
For the prediction-selection loop, we set $K$ to 2. For the LLM, we use LLaMA 3 70B \cite{llama3modelcard} and vLLM \cite{kwon2023efficient} as the inference software. 

\subsection{Main Results}
The performance of \method{} in comparison with prior methods is shown in Table \ref{tab:main}. 
According to the metrics in the table, we have several observations:

Firstly, the proposed \method{} demonstrates consistent improvement in all three datasets, showing the effectiveness of the proposed framework that adopts a dual-branch traffic flow predictor and an LLM-based selector.
Smaller models often fail to provide satisfactory predictions under test-time distribution changes, as they fall short in reason and generalization, leaving room for improvement.

Secondly, the methods that utilize the generative ability of LLMs (\emph{i.e.} LLM-MPE) do not perform well on all datasets. As we can see on the PEMS03 and PEMS04 datasets, their predictions are generally worse or similar compared to simple methods using graph neural networks (\emph{e.g.} STSGCN, COOL). Since LLMs are not adept at capturing complex spatio-temporal relations, it is reasonable that we see such results on datasets with larger networks like PEMS03.

Thirdly, the proposed \method{} generally performs better than graph-based methods (\emph{e.g.} ASTGCN, STSGCN) and hypergraph-based methods (\emph{e.g.} HGCN, DyHSL), which shows that our method has the potential to take advantage of both complex spatio-temporal relations captured by the predictor and the knowledge of large language models, achieving LLM enhanced traffic flow forecasting.

\input{Tab/ablation}

\subsection{Ablation Study}
We also perform ablation studies on the PEMS08 dataset, and the results are shown in Table \ref{tab:abl}. Specifically, we perform the following experiments. \textbf{E1} measures the performance of the graph branch only without the LLM-based selector. \textbf{E2} measures the performance of the hypergraph branch without the selector. The vanilla version (E1 and E2) of both branches performs much worse than \method{}. \textbf{E3} uses the graph branch in conjunction with the selector, which leads to performance degradation compared to \method{}. This suggests that non-pair-wise relations are important in traffic flow forecasting. \textbf{E4} only uses the hypergraph branch together with the selector. Similarly, it performs worse than \method{}, which shows that pair-wise and non-pair-wise relations are both important. Moreover, E3 and E4 perform better than E1 and E2, which shows the effect of the LLM-based selector. \textbf{E5} removes the transformations, \emph{i.e.} $\mathcal T = \emptyset$. This leads to slightly worse performance, showing the effectiveness of providing more choices. \textbf{E6} removes the ranking loss, which means that the predictor is not further trained with the results from the selector (reducing to $K=1$). This experiment demonstrates the effectiveness of supervising the predictor with a ranking loss using the selection results.

\begin{figure}
    \centering
    \includegraphics[width=\linewidth]{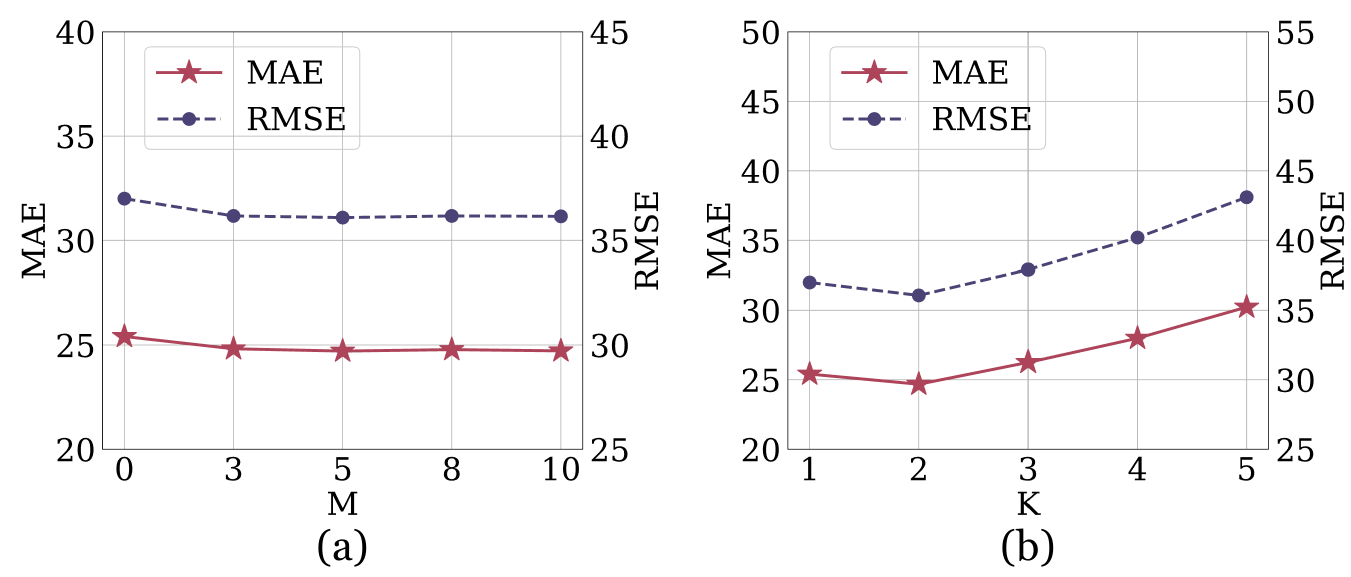}
    \vspace{-5mm}
    \caption{The forecasting errors under different hyper-parameters, \emph{i.e.} $M$s (left) and $K$s (right).}
    \label{fig:sensitivity}
\end{figure}
\subsection{Hyper-parameter Analysis}
We perform experiments with respect to two hyper-parameters: \textbf{(a)} $M$, \emph{i.e.} the number of iterations when training with the ranking loss in Eq. \ref{eq:metric} and Eq. \ref{eq:loss}, and \textbf{(b)} the number of $K$, \emph{i.e.} the number of prediction-selection iterations in Algorithm \ref{alg:main}. The results on the PEMS08 dataset are shown in Figure \ref{fig:sensitivity}. As can be seen from the figure, when $M$ increases, both MAE and RMSE decrease, and then plateau after around 5. This shows that the predictor converges, and therefore, we set $M$ to 5. For another hyper-parameter $K$, the optimal performance is achieved when $K$ is set to 2, and with more iterations, the error increases. One reason is that we use the same prompt across iterations without contexts of previous selections to save computation, so the same factor may be considered multiple times, leading to decreased accuracy. In practice, $K=2$ is good enough without too much computation for the LLM-based selector.

\begin{figure}[ht]
    \centering
    \includegraphics[width=\linewidth]{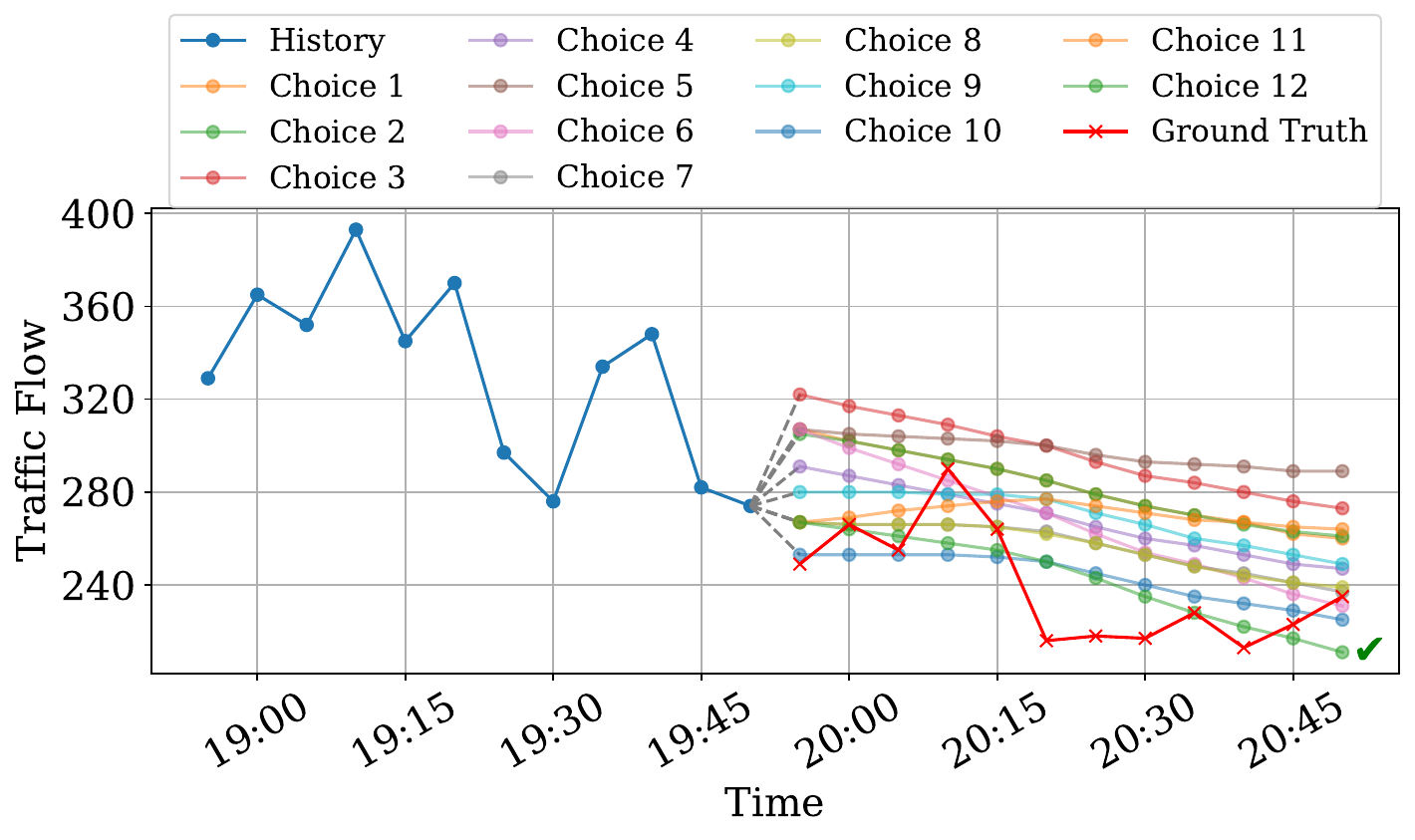}
    \vspace{-5mm}
    \caption{Visualization of historical data, choices, and the prediction in different time periods. The selected choice is marked with a green tick.}
    \label{fig:case}
\end{figure}

\begin{figure}[ht]
    \centering
    \includegraphics[width=\linewidth]{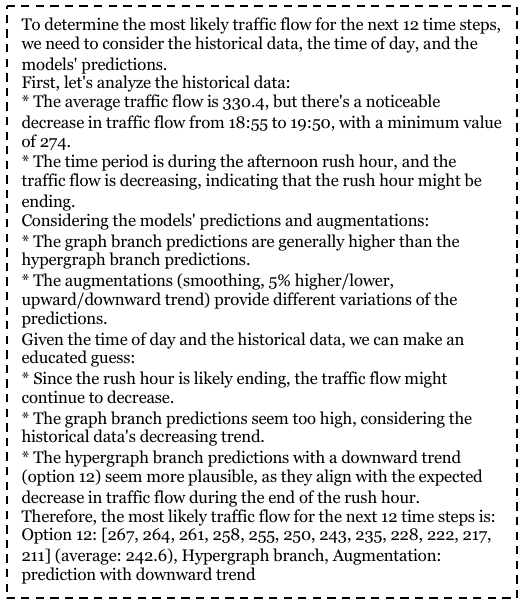}
    \vspace{-5mm}
    \caption{The LLM's analysis when selecting data.}
    \label{fig:analysis}
\end{figure}

\begin{figure}[th]
    \centering
    \includegraphics[width=\linewidth]{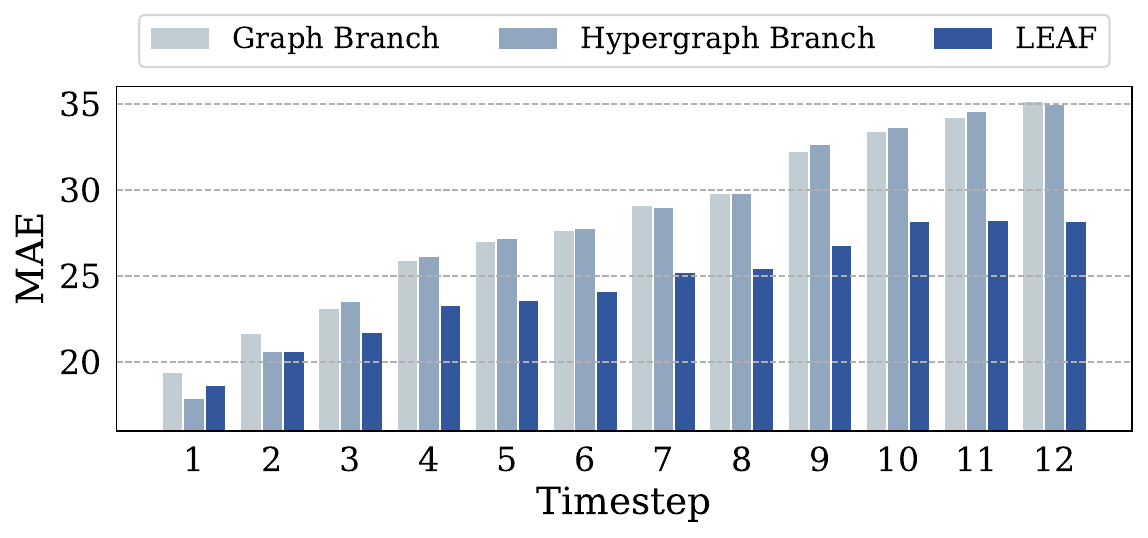}
    \vspace{-7mm}
    \caption{The Mean Absolute Error (MAE) under different timesteps. \method{} better reduces long-term errors.}
    \label{fig:mae}
\end{figure}

\begin{figure*}[t]
    \centering
    \includegraphics[width=\linewidth]{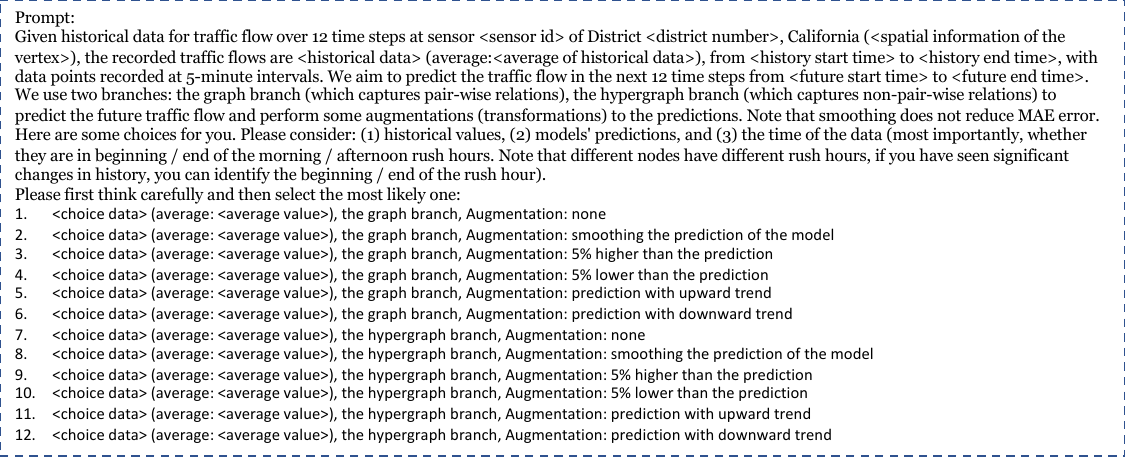}
    \caption{The details about the prompt.}
    \label{fig:exampleprompt}
  \vspace{-1mm}
\end{figure*}

\subsection{Visualization}
\noindent\textbf{Visualization of the Selection}.
We first provide visualizations of the historical traffic flow data, the choices in $\mathcal C_i$, and the option selected by the LLM-based selector (marked with a green tick). The results on the PEMS03 dataset are shown in Figure \ref{fig:case}. Moreover, we also provide the analysis of the LLM-based selector, which is shown in Figure \ref{fig:analysis}.
According to the results, we can see that the traffic flow is generally going downward, since it is the end of the rush hour. In Figure \ref{fig:case}, we can see that the LLM-based selector chooses the option with the lowest traffic flows. From its analysis in Figure \ref{fig:analysis}, we can see that it understands that the time period to forecast is around the end of the rush hours, which is the reason why it selects the lowest option. This suggests that the LLM-based selector is able to understand changing traffic conditions.

\smallskip
\noindent\textbf{Visualization of Errors in Different Timesteps}. We then provide visualization of the mean absolute error (MAE) under different forecasting timesteps in Figure \ref{fig:mae}. The experiments are performed on the PEMS03 dataset, where we compare the MAE values of our framework to its two branches (\emph{i.e.} the graph branch and the hypergraph branch). The results show that \method{} reduces forecasting errors generally. Although the errors are similar in the first few timesteps, \method{} quickly diverges from the two branches, resulting in significantly lower errors in the long term. This suggests that with the help of the discriminative ability of the LLM-based selector, our method can select predictions that are more accurate in the long run.

\smallskip
\noindent\textbf{Visualization of the Prompt.} We provide an example of the prompt in Figure \ref{fig:exampleprompt}. The prompt contains general information about the task and the traffic data, spatio-temporal information, the historical data, the task instructions, and the choice set constructed in Eq. \ref{eq:choice}. Additionally, we also describe the transformations applied to the prediction of the dual-branch predictor. These sources of information will be helpful for the LLM to decide the most appropriate choice.

%% file: Tab/main_old.tex
{
\begin{table*}[t]
    \centering
    \setlength{\tabcolsep}{6pt}
    \resizebox{\textwidth}{!}{%
    \begin{tabular}{l ccc c ccc c ccc}
    \toprule
    \multirow{2}{*}{\textbf{Model}}  & \multicolumn{3}{c}{\textbf{PEMS03}}    && \multicolumn{3}{c}{\textbf{PEMS04}}      && \multicolumn{3}{c}{\textbf{PEMS08}}\\\cmidrule{2-4} \cmidrule{6-8} \cmidrule{10-12}
    & MAE & RMSE & MAPE             && MAE & RMSE & MAPE               && MAE & RMSE & MAPE             \\ \midrule
    DCRNN \cite{li2018diffusionDCRNN} & 29.99 & 39.52 & 21.33  && 34.36 & 46.19  & 24.73 && 31.41 & 43.91 & 15.44 \\
    ASTGCN \cite{guo2019attention}     & 28.4  & 41.94 & 15.78  && 33.09 & 46.08  & 18.19 && 29.20  & 41.16 & 12.76 \\
    STSGNN \cite{song2020spatialSTSGCN}     & 28.21 & 43.43 & 15.49  && 33.43 & 45.69  & 18.89 && 29.58 & 41.95 & 12.90  \\
    HGCN \cite{wang2022multitask}       & 28.43 & 43.92 & 15.39  && 35.77 & 49.92  & 20.11 && 28.83 & 39.65 & 12.63 \\
    DyHSL \cite{zhao2023dynamic}      & 27.10  & 41.59 & 14.31  && 33.36 & 46.96  & 19.64 && 27.34 & 39.05 & 11.56 \\
    STAEformer \cite{liu2023spatioSTAEformer} & 27.87 & 37.31 & 16.49  && 33.77 & 45.50   & 18.36 && 27.43 & 38.16 & 11.36 \\
    COOL  \cite{ju2024cool}      & 27.51 & 41.11 & 14.73  && 34.68 & 47.22  & 19.72 && 27.22 & 38.47 & 11.72 \\
    LLM-MPE \cite{liang2024exploring}    & 33.82 & 47.06 & 20.40   && 35.63 & 51.41  & 18.19 && 26.42 & 40.02 & 10.61 \\\midrule
    \rowcolor{gray!10}\textbf{\method{}} & \textbf{25.46} & \textbf{35.17} & \textbf{14.22} && \textbf{31.49} & \textbf{44.45} & \textbf{17.53} && \textbf{24.68} & \textbf{36.07} & \textbf{10.56} \\
    \bottomrule
\end{tabular}
    }%
\caption{Forecasting errors on PEMS03, PEMS04, and PEMS08 datasets.}
\label{tab:main}
\vspace{-2mm}
\end{table*}

}

%% file: Tab/ablation.tex
{
\begin{table}[tb]
\centering
\resizebox{0.47\textwidth}{!}{
\begin{tabular}{lccc}
\toprule[1pt]
\bf{Experiments} & MAE & RMSE & MAPE \\\midrule
E1: Graph branch & 29.12 & 41.36 & 13.54 \\\midrule
E2: Hypergraph branch & 27.94 & 39.11 & 11.82 \\\midrule
E3: \emph{w/o} hypergraph branch & 26.29 & 38.18 & 12.83 \\\midrule
E4: \emph{w/o} graph branch & 25.80 & 37.23 & 11.00 \\\midrule
E5: \emph{w/o} transformation & 25.47 & 36.47 & 11.01 \\\midrule
E6: \emph{w/o} ranking loss & 25.41 & 37.00 & 11.34 \\\midrule
    \rowcolor{gray!10} \textbf{LEAF} & \textbf{24.68} & \textbf{36.07} & \textbf{10.56} \\
\bottomrule[1pt]
\end{tabular}}
\caption{Ablation study on the PEMS08 dataset.}
\label{tab:abl}
\vspace{-3mm}
\end{table}
}

%% file: 6_conclusion.tex
\section{Conclusion}
In this paper, we propose a novel framework named Large \underline{L}anguage Model \underline{E}nhanced Tr\underline{a}ffic \underline{F}low Predictor (\method{}), consisting of a dual-branch traffic flow predictor and an LLM-based selector. The predictor adopts two branches: the graph branch and the hypergraph branch, capturing the pair-wise and non-pair-wise relations, respectively. The selector uses the discriminative ability of the LLM to choose the best forecast of the predictor. The selection results are then used to supervise the predictor. We perform extensive experiments to demonstrate the effectiveness of \method{}.

\section*{Limitations}
One limitation of this work is that we only focus on the traffic flow forecasting domain, due to the scope of this paper and the limited computational resources. However, this framework can be extended to more generalized spatio-temporal forecasting problems. Besides, we have not evaluated our methods on other traffic flow forecasting datasets due to limited resources. 

Another limitation is that the LLM is not fine-tuned during the process. The LLM selector in the LEAF framework can be further optimized throughout the process. Since the traffic datasets are relatively small, a potential solution is to use parameter-efficient fine-tuning strategies, including LoRA, adapter layers, or prefix-tuning to enhance the performance of the LLM selector. More specifically, given the final prediction results from the iterative prediction-selection cycles, the LLM is then fine-tuned with the answer (in the choices) nearest to the final prediction. This will potentially enhance the ability of the LLM selector.

\vspace{0.5mm}
\section*{Acknowledgment}
\vspace{0.5mm}
This paper is partially supported by grants from the National Key Research and Development Program of China with Grant No. 2023YFC3341203 and the National Natural Science Foundation of China (NSFC Grant Number 62276002). The authors are grateful to the anonymous reviewers for critically reading this article and for giving important suggestions to improve this article.